\documentclass[runningheads]{llncs}
\usepackage{graphicx}
\usepackage{newfloat}
\usepackage{listings}
\usepackage{amsmath,amssymb,bm,pifont}
\usepackage{color}
\usepackage{multirow}
\usepackage{soul}
\usepackage{booktabs}
\usepackage{bigstrut}
\usepackage[misc]{ifsym}
\usepackage{makecell}
 \usepackage{bigstrut}
\usepackage[table]{xcolor}
\begin{document}
\title{Source-Free Domain Adaptation  for Medical Image Segmentation via Prototype-Anchored Feature Alignment and Contrastive Learning}
\authorrunning{Q. Yu et al.}

\author{Qinji Yu\inst{1} \and
Nan Xi\inst{2} \and
Junsong Yuan\inst{2} \and Ziyu Zhou\inst{1} \and Kang Dang\inst{3 \thanks{Co-Corresponding Author}}\and Xiaowei Ding\inst{1, 3\footnotemark[1]} \textsuperscript{(\Letter)}}

\institute{Shanghai Jiao Tong University, Shanghai, China \and
State University of New York at Buffalo, New York, USA \and
VoxelCloud, Inc., Los Angeles, USA \\
\email{dingxiaowei@sjtu.edu.cn}}

\maketitle              
\begin{abstract}
Unsupervised domain adaptation (UDA) has increasingly gained interests for its capacity to transfer the knowledge learned from a labeled source domain to an unlabeled target domain. However, typical UDA methods require concurrent access to both the source and target domain data, which largely limits its application in medical scenarios where source data is often unavailable due to privacy concern. To tackle the source data-absent problem, we present a novel two-stage source-free domain adaptation (SFDA) framework for medical image segmentation, where only a well-trained source segmentation model and unlabeled target data are available during domain adaptation. Specifically, in the prototype-anchored feature alignment stage, we first utilize the weights of the pre-trained pixel-wise classifier as source prototypes, which preserve the information of source features. Then, we introduce the bi-directional transport to align the target features with class prototypes by minimizing its expected cost. On top of that, a contrastive learning stage is further devised to utilize those pixels with unreliable predictions for a more compact target feature distribution.  Extensive experiments on a cross-modality medical segmentation task demonstrate the superiority of our method in large domain discrepancy settings compared with the state-of-the-art SFDA approaches and even some UDA methods. Code is available at: \url{https://github.com/CSCYQJ/MICCAI23-ProtoContra-SFDA}.

\end{abstract}
\section{Introduction}
The inception of deep neural networks has revolutionized the landscape of medical image segmentation \cite{ronneberger2015u,zhou2018unet++}. This tremendous success, however, is conditioned on the assumption that the training and testing data are drawn from the same distribution. Unfortunately, in real-world clinical scenarios, due to different acquisition protocols or various imaging modalities, domain shift is widespread between training (\textit{i.e.}, source domain) and testing (\textit{i.e.}, target domain) datasets \cite{stan2021privacy}. This distribution gap usually degenerates the model performance on the target domain. To achieve reliable performance across different domains, a straightforward way is manually labeling some target data and fine-tuning the pre-trained model on them \cite{motiian2017few}. However, obtaining expert-level annotation data in the medical imaging domain incurs significant time and expense \cite{yu2021location}. Recently, unsupervised domain adaptation (UDA) has been widely investigated to reduce domain gap through transferring the knowledge learned from a rich-labeled source domain to an unlabeled target domain \cite{tzeng2017adversarial,dou2018unsupervised,chen2020unsupervised,xian2023unsupervised}. Existing UDA methods typically require sharing source data during adaptation, and enforce distribution alignment to diminish the domain discrepancy between source and target domains. This requirement limits the application of UDA methods when source domain data are not accessible. Hence, some very recent works have started to explore a more pratical setting, source-free domain adaptation (SFDA), that adapts a pre-trained source model to unlabeled target domains without accessing any source data \cite{chen2021source,ding2022source,yang2022source,liu2022source,bateson2022source,xu2022denoising}.

\begin{figure}[tb]
\includegraphics[width=\textwidth]{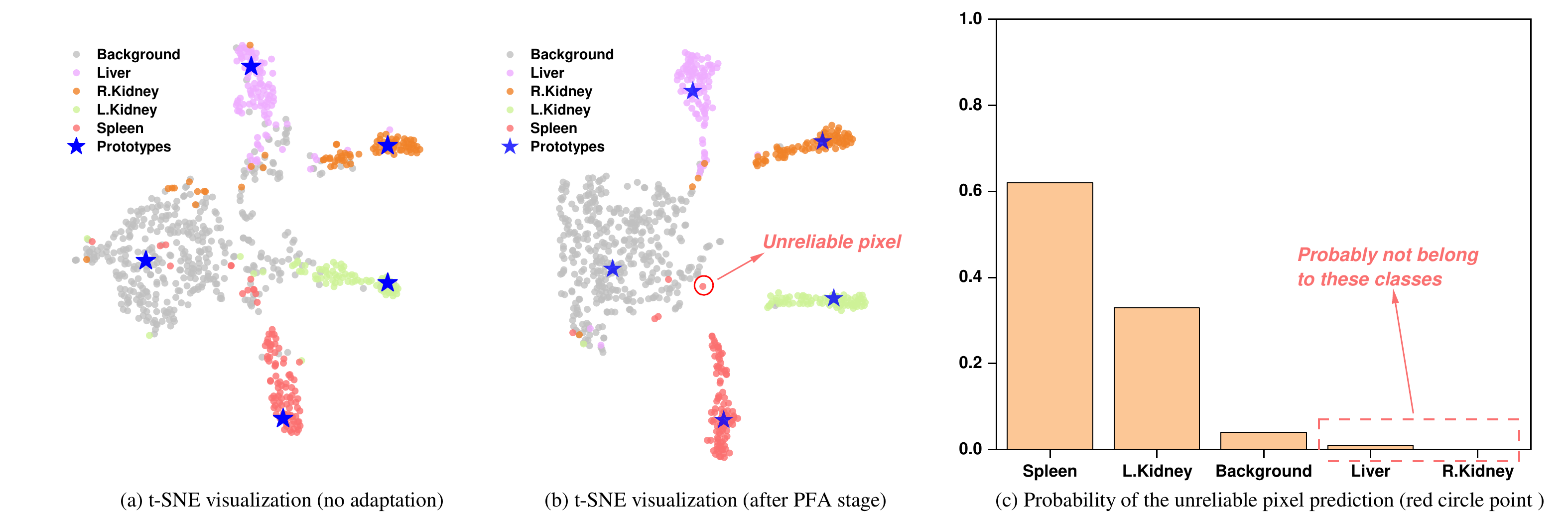}
\caption{(a→b) t-SNE visualization of target feature distributions in embedding space before and after Prototype-anchored Feature Alignment (PFA). (c) Category-wise probability of the unreliable pixel in (b).} \label{fig1}
\end{figure}

Among these methods, \cite{chen2021source} and \cite{xu2022denoising} focus on generating reliable pseudo labels for target domain data by developing various denoising strategies. Unavoidably, these self-training methods depends heavily on initial probability maps produced by the source model, which are considerably unreliable when the domain discrepancy is large (\textit{e.g.}, CT and MRI). To relieve the issues caused by noisy pseudo labels, Bateson et al. \cite{bateson2022source} proposed a prior-aware entropy minimization method to minimize the label-free entropy loss for target predictions. Furthermore, unlike the above self-adaption methods, Yang et al. \cite{yang2022source} utilized the statistic information stored in the batch normalization layer of the source model and mutual Fourier Transform to synthesize the source-like image. However, the quality of the generated image is still influenced by the domain discrepancy.

In this work, we propose a novel SFDA framework for cross-modality medical image segmentation. Our framework contains two sequentially conducted stages, \textit{i.e.}, Prototype-anchored Feature Alignment (PFA) stage and Contrastive Learning (CL) stage. As previous works \cite{liu2022source} noted, the weights of the pre-trained classifier (\textit{i.e.}, projection head) can be employed as the source prototypes during domain adaptation. That means we can characterize the features of each class with a source prototype and align the target features with them instead of the inaccessible source features. 
To that end, during the PFA stage, we first provide a target-to-prototype transport to ensure the target features get close to the corresponding prototypes. Then, considering the trivial solution that all target features are assigned to the dominant class prototype (\textit{e.g.}, background), we add a reverse prototype-to-target transport to encourage diversity. However, although most target features have been assigned to the correct class prototype after PFA, some hard samples with high prediction uncertainty still exist in the decision boundary (see Fig. \ref{fig1}(a→b)). Moreover, we observe that those unreliable predictions usually get confused among only a few classes instead of all classes \cite{wang2022semi}. Taking the unreliable pixel in Fig. \ref{fig1}(b,c) for example, though it achieves similar high probabilities on the spleen and left kidney, the model is pretty sure about this pixel not belonging to the liver and right kidney. Inspired by this, we use confusing pixels as the negative samples for those unlikely classes, and then introduce the CL stage to pursue a more compact target feature distribution. Finally, we conduct experiments on a cross-modality abdominal multi-organ segmentation task. With only a source model and unlabeled target data, our method outperforms the state-of-the-art SFDA and even achieves comparable results with some classical UDA approaches.

\section{Methods}
We are first provided a segmentation model $\mathcal{M}^{s}$ trained on $N_s$ labeled samples $\left\{(x_{n}^{s}, y_{n}^{s})\right\}_{n=1}^{N_{s}}$ from the source domain $\mathcal{D}^{s}$, and an unlabeled dataset with $N_t$ samples $\left\{x_{m}^{t}\right\}_{m=1}^{N_{t}}$ from the target domain $\mathcal{D}^{t}$, where $x^s, x^t \in \mathbb{R}^{H \times W \times D}$, $y_{n}^{s} \in \mathbb{R}^{H \times W}$, $H$ and $W$ are the height and width of the samples. The goal of SFDA is to adapt the source model $\mathcal{M}^{s}$ with only unlabeled $x^t$ to predict pixel-wise label $y^t$ for the target domain data. In general, the segmentation model consists of two parts: 1) a feature extractor $F_{\theta}:x_i\to \bm{f}_i\in\mathbb{R}^{D_f}$, parameterized by $\theta$, mapping each pixel $i \in \{1,\cdots,H \times W\}$ in image $x$ to the feature $\bm{f}_i$ in the embedding space; 2) a one-layer pixel-wise classifier $\phi:\bm{f}_i\to \bm{p}_i \in \mathbb{R}^{C}$, that projects pixel feature into the semantic label space with $C$ classes. 

\begin{figure}[h]
\includegraphics[width=\textwidth]{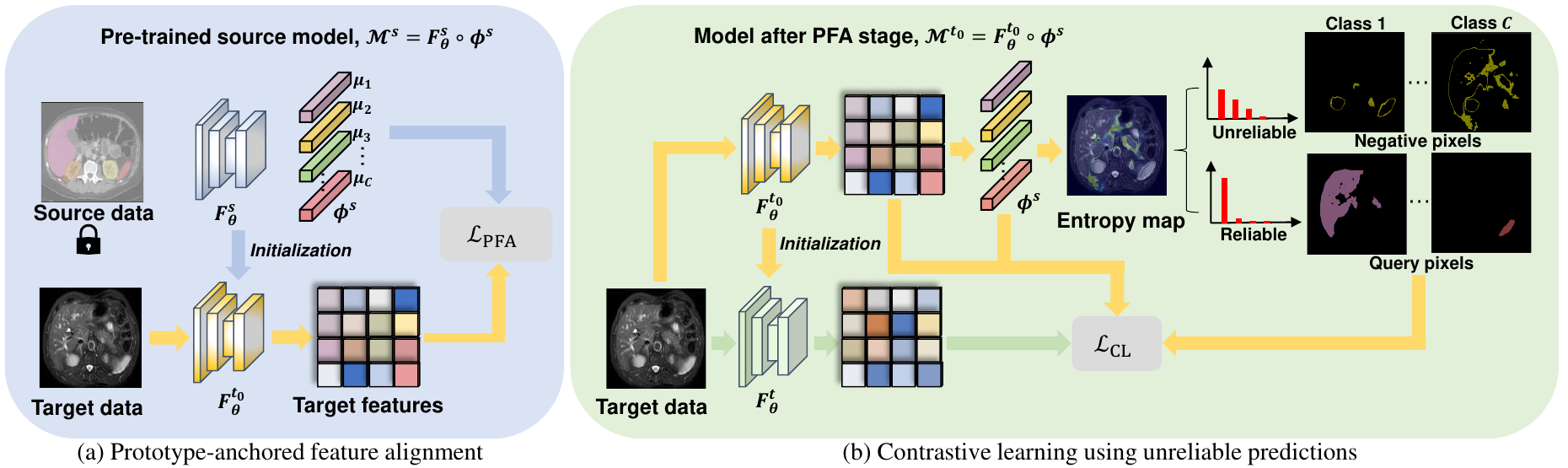}
\caption{An overview of the proposed two-stage SFDA framework. (a) is the first PFA stage. We freeze the classifier $\phi^s$ and use its weights for prototype-anchored feature alignment. (b) is the following CL stage. Given a target image, we first use $\mathcal{M}^{t_0}$ to make a prediction, and seperate the pixels into query one and negative ones for each class based on their 
reliability (entropy). Then, features of query pixels come from $F^{t}_{\theta}$ (query samples), while features of negative pixels are from $F^{t_0}_{\theta}$ (negative samples), when minimizing $\mathcal{L}_{\mathrm{CL}}$. } \label{fig2}
\end{figure}

In the SFDA task, the source classifier $\phi^{s}$  encounters a domain shift problem when classifying the target domain feature. To tackle this challenge, we propose a novel SFDA framework mainly including two stages, shown in Fig. \ref{fig2}. We will elaborate on the details in the following.

\subsection{Prototype-Anchored Feature Alignment}
Since source data is not available, explicit feature alignment that directly minimizes the domain gap between the source and target data like many UDA methods \cite{chen2020unsupervised,han2021deep} is inoperative. As shown by previous methods \cite{liu2022source},  the weights $[\bm{\mu}_1,\bm{\mu}_2,\cdots,\bm{\mu}_C] \in \mathbb{R}^{D_f \times C}$ of the source domain classifier $\phi^{s}$ can be interpreted as the source prototypes, which characterize the features of each class. Thus, we introduce a bi-directional transport cost to align the target features with these prototypes instead of the unaccessible source features.


Following \cite{zheng2021exploiting}, given a mini-batch $\left\{x_{m}^{t}\right\}_{m=1}^{M}$ with $M$ images, we first adopt the cosine distance $d(\bm{\mu}_{c},\bm{f}_{m, i}^{t})= 1-\langle\bm{\mu}_{c},\bm{f}_{m, i}^{t}\rangle$ to define a point-to-point transport cost between $\bm{f}_{m, i}^{t}$ and $\bm{\mu}_{c}$, where $\langle\cdot,\cdot\rangle$ is the cosine similarity. Then, a conditional distribution $\pi_{\theta}\left(\bm{\mu}_{c} \mid \bm{f}_{m, i}^{t}\right)$ specifying the probability of transporting from $\bm{f}_{m, i}^{t}$ to $\bm{\mu}_{c}$ can be constructed as,
\begin{equation}
    \pi_{\theta}\left(\bm{\mu}_{c} \mid \bm{f}_{m, i}^{t}\right)=\frac{\hat{p}\left(\bm{\mu}_{c}\right) \exp \left(\bm{\mu}_{c}^{T} \bm{f}_{m, i}^{t} / \tau \right)}{\sum_{c^{\prime}=1}^{C} \hat{p}\left(\bm{\mu}_{c^{\prime}}\right) \exp \left(\bm{\mu}_{c^{\prime}}^{T} \bm{f}_{m, i}^{t} / \tau \right)}
\end{equation}
where $\tau$ is the temperature parameter, and $\hat{p}\left(\bm{\mu}_{c}\right)$ is the prior distribution (\textit{i.e.}, class propotion) over the $C$ classes for the target domain. As the true class distribution is unavailable in the target domain, we use the EM algorithm to infer $\hat{p}\left(\bm{\mu}_{c}\right)$ instead of using a uniform prior distribution (see more details in \cite{tanwisuth2021prototype}). Note that in Eq. 1, a target point is more likely to be
transported to the class prototypes closer to it or those with higher class propotion.

With the conditional distribution and point-to-point transport cost, we can derive the target-to-prototype (T2P) expected cost of moving the target features in this mini-batch to source prototypes,
\begin{equation}
    \mathcal{L}_{\mathrm{T2P}} = \frac{1}{M\times H \times W} \sum_{m=1}^{M} \sum_{i=1}^{H \times W} \sum_{c=1}^{C} d(\bm{\mu}_{c},\bm{f}_{m, i}^{t}) \pi_{\theta}\left(\bm{\mu}_{c} \mid \bm{f}_{m, i}^{t}\right)
\end{equation}

In this target-to-prototype direction, we assign each target pixel to the prototypes according to their similarities and the class distribution. However, like many entropy minimization methods \cite{bateson2022source,bateson2022test}, optimizing target-to-prototype cost alone may result in degenerate trivial solutions, biasing the prediction towards a single dominant class \cite{tanwisuth2021prototype}. To avoid mapping most of the target features to only a few
prototypes, we add a prototype-to-target (P2T) transport cost in the opposite direction, which ensures that each prototype can be assigned to some target features. Similarly, we have:
\begin{equation}
    \mathcal{L}_{\mathrm{P2T}} = \sum_{c=1}^{C} \hat{p}\left(\bm{\mu}_{c}\right) \sum_{m=1}^{M} \sum_{i=1}^{H \times W}d(\bm{\mu}_{c},\bm{f}_{m, i}^{t}) \frac{\exp \left(\bm{\mu}_{c}^{T} \bm{f}_{m, i}^{t} / \tau\right)}{\sum_{m^{\prime}=1}^{M} \sum_{i^{\prime}=1}^{H \times W} \exp \left(\bm{\mu}_{c}^{T} \bm{f}_{m^{\prime}, i^{\prime}}^{t} / \tau\right)} 
\end{equation}

Then, combining the conditional transport cost in these two directions, we define the total prototype-anchored feature alignment (PFA) loss:
\begin{equation}
    \mathcal{L}_{\mathrm{PFA}} = \mathcal{L}_{\mathrm{T2P}}+\mathcal{L}_{\mathrm{P2T}}
\end{equation}

Similar to \cite{ding2022source}, we initialize the adaptation model $\mathcal{M}^{t_0}$ with the pre-trained source model $\mathcal{M}^{s}$ and fix the weights of the classifier during adaptation.

\subsection{Contrastive Learning using Unreliable Predictions}

After the PFA stage, the clusters of target features are shifted towards their corresponding source prototypes, which brings remarkable improvements for the initial noisy prediction (see Fig.\ref{fig3}(b)). To further improve the compactness of the target feature distribution, previous self-training methods mainly focus on strengthening the reliability of pseudo labels by developing denoising strategies \cite{chen2021source,xu2022denoising}, but discard those low-confidence predictions. However, such contempt for unreliable predictions may result in information loss. For example, in Fig.\ref{fig1}(c), the probability of the unreliable pixel hovers between spleen and left kidney, yet is confident enough to indicate the categories it does not belong to.

With this intuition, we denote $\bm{p}_{m,i}^{t}$ as the softmax probabilities generated by model $\mathcal{M}^{t_0}$ for the target data $x_{m,i}^{t}$. Then, for each class $c$, we construct three components, named query samples, positive prototypes, and negative samples, to explore those unreliable predictions as \cite{wang2022semi}.
\\
\textit{\textbf{Query Samples.}} During training, we employ the per-pixel entropy as uncertainty metric \cite{wang2022semi}, and sample the pixels with low entropy (reliable pixel) in the current mini-batch as query candidates. We denote the set of features of all query pixels for class $c$ as $\mathcal{P}_{c}$,
\begin{equation}
    \mathcal{P}_{c} = \{ \bm{f}_{m, i}^{t} \ \vert \ \mathcal{H}({\bm{p}_{m,i}^{t}}) \leq \gamma_c, \ \arg\max\limits_{c^{\prime}}{\bm{p}_{m,i}^{t}}=c
     \}
\end{equation}
where $\mathcal{H}(\cdot)$ is the entropy of the input probabilities and $\gamma_c$ is the entropy threshold for class $c$. Here we set $\gamma_c$ as the $\alpha_c$-th percentile of all the entropy values of pixels assigned a pseudo label $c$.\\
\textit{\textbf{Positive Prototypes.}} The positive prototype is the same for all query pixels from the same class. Instead of using the center of query samples like \cite{wang2022semi}, we set them the same as the previous source prototype, which is denoted as $\bm{z}_c^{+}=\bm{\mu}_c$.\\
\textit{\textbf{Negative Samples.}} For a query sample from class $c$, its qualified negative samples should satisfy: 1) unreliable; 2) highly probable not belong to class $c$. Therefore, we introduce the pixel-level category order $\mathcal{O}_{m, i}^{t} = \mathrm{argsort}({\bm{p}_{m,i}^{t}})$. For example, we have $\mathcal{O}_{m, i}^{t}(\arg\max{\bm{p}_{m,i}^{t}})=1$ and $\mathcal{O}_{m, i}^{t}(\arg\min{\bm{p}_{m,i}^{t}})=C$. Thus, we can use $\mathcal{O}_{m, i}^{t}(c)$ to define the set of all negative samples:
\begin{equation}
    \mathcal{N}_{c} = \{ \bm{f}_{m, i}^{t} \ \vert \ \mathcal{H}({\bm{p}_{m,i}^{t}})>\gamma_c, \
    \mathcal{O}_{m, i}^{t}(c) \geq r_l \}
\end{equation}
where $r_l$ is the low rank threshold and is set to 3 in our task.

With the above definition, we have the pixel-level contrastive loss as:
\begin{equation}
\mathcal{L}_{\mathrm{CL}}=  -\frac{1}{C \times K} \sum_{c=1}^C \sum_{k=1}^K \log \left[\frac{e^{\left\langle\bm{z}_{c,k}, \bm{z}_{c}^{+}\right\rangle / \tau}}{e^{\left\langle\bm{z}_{c,k}, \bm{z}_{c}^{+}\right\rangle / \tau}+\sum_{j=1}^N e^{\left\langle\bm{z}_{c,k}, \bm{z}_{c,k,j}^{-}\right\rangle / \tau}}\right]
\end{equation}
where $K$ is the number of query samples, and $\bm{z}_{c,k} \in \mathcal{P}_{c}$ denotes the $k$-th query sample from class $c$. Each query sample is paired with a positive prototype $\bm{z}_{c}^{+}$ and $N$ negative samples $\bm{z}_{c,k,j}^{-} \in \mathcal{N}_{c}$. 

\section{Experiments and Results}
\subsection{Experimental Setup}
\textit{\textbf{Datasets and Evaluation Metrics.}} We evaluate our SFDA approach on a cross-modality abdominal multi-organ segmentation task. For the abdominal datasets, we obtain 20 MRI volumes from the 2019 CHAOS Challenge \cite{kavur2021chaos} and 30 CT volumes from MICCAI 2015 \cite{landman2015miccai}, respectively. Both datasets are under the Creative Commons Attribution 4.0 International license and involve segmentation masks for the following abdominal organs: liver, right kidney, left kidney and spleen. We complete adaptation experiments both in the ``MRI to CT'' direction and in the ``CT to MRI'' direction. For the ``MRI to CT'' direction, we take the MRI modality to train the source model and vice verse. Both modalities are randomly divided into 80\% for domain adaptation training and 20\% for evaluation. For both datasets, we discard the axial slices that do not contain foreground and crop out the non-body region \cite{bian2021domain}. The value range in CT volumes is first clipped to $[-125,275]$. Then min-max normalization has been performed on both datasets to normalize the intensity value to $[0,1]$. After that, all the MRI and CT volumes are uniformly resized to $256\times 256$ in axial plane. Due to the large variance in the slice thickness of CT and MRI modality, we split the volume into slices for the model training.

\begin{figure}[b]
\includegraphics[width=\textwidth]{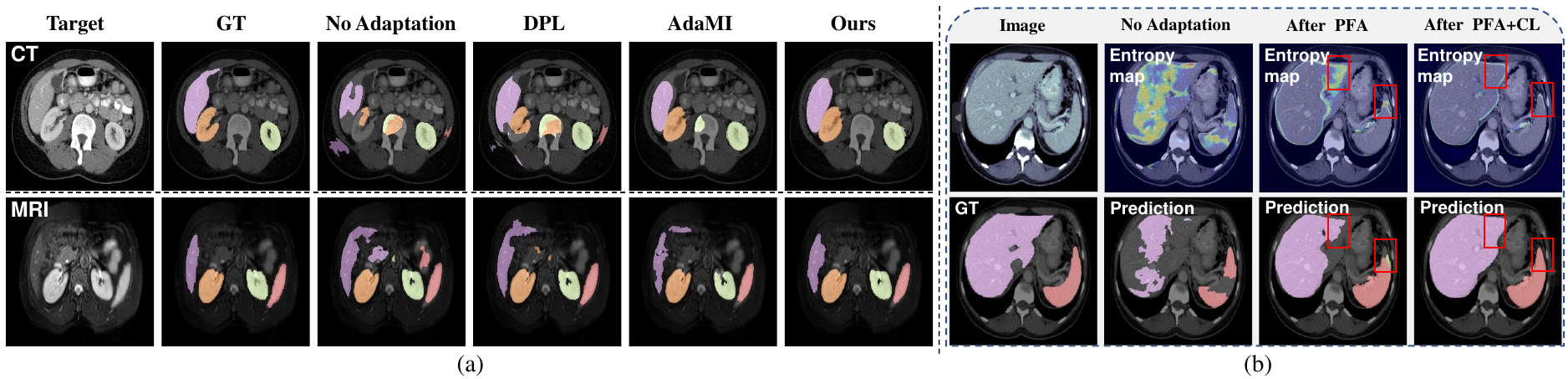}
\caption{(a) Qualitative segmentation results of different methods for abdominal images. (b) Visualized evolution of the model uncertainty and predictions in different stages.} \label{fig3}
\end{figure}

For the evaluation, two main metrics, dice similarity coefficient (Dice)  and average symmetric surface distance (ASSD) are used to quantitatively evaluate the segmentation results \cite{chen2020unsupervised,stan2021privacy}.\\
\textit{\textbf{Implementation Details.}} We adopt classic U-Net structure for the segmentation model as the previous work \cite{bateson2022source}. The source segmentation model is trained in a fully-supervised manner for 10k iterations. During adaptation, we use Adam optimizer with the learning rate $1 \times 10^{-4}$ and a weight decay of $5 \times 10^{-4}$.  The temperature $\tau$ and batch size is set as 0.1 and 16, respectively. In PFA stage, we freeze the classifier and optimize $F_\theta^{t_0}$ for 200 iterations. In CL stage, we empirically set hyper-parameters $\alpha_c=80$, $K=64$, and $N=256$ for all classes.  All experiments are conducted with PyTorch on a single NVIDIA RTX 3090 GPU of 24 GB memory. Data augmentation such as random cropping, rotation, and brightness are adopted for source domain training and target domain adaptation.

\subsection{Results of Source-free Domain Adaptation}
\textit{\textbf{Comparision with Other Methods.}} In our experiments, ``no adaptation'' lower bound denotes learning a model on the source domain and directly test on the target domain without adaptation. And ``supervised'' upper bound means training and testing in the same target domain. We compared our methods with recent SFDA methods all designed for medical image segmentation scenarios, including a denoised pseudo-labeling approach (DPL) \cite{chen2021source}, a prior-aware entropy minimization approach (AdaMI) \cite{bateson2022source}, a fourier style mining approach (FSM) \cite{yang2022source}, and a feature map statistics-guided approach \cite{hong2022source}. We also considered top-performing UDA methods (\textit{i.e.}, SIFA \cite{chen2020unsupervised}, DAG-Net \cite{xian2023unsupervised}). For a fair comparison, we utilized the same backbone for these methods \cite{chen2021source,bateson2022source,yang2022source,chen2020unsupervised} and reimplemented them according to their official codes. Note that we reported the results of methods \cite{hong2022source,xian2023unsupervised} from papers, since their official codes were not released.

\begin{table*}[tb]
    \renewcommand\arraystretch{1.0}
    \centering
        \caption{Comparision with other methods on abdominal multi-organ datasets.}
        \resizebox{1.0\textwidth}{!}{%
        \setlength\tabcolsep{3.0pt}
        \scalebox{1.00}{
        \begin{tabular}{c c|c  c  c  c c ||c  c  c  c c }
    
            \hline
            &\multicolumn{9}{c}{\textbf{Abdominal MRI $\rightarrow$ Abdominal CT}}\\
            \hline
            \multirow{2}{*}{Method} &\multirow{2}{*}{Source-free}  &\multicolumn{5}{c||}{\textbf{Dice (\%, mean$\pm$std)~$\uparrow$ }} &\multicolumn{5}{c}{\textbf{ASSD (voxel, mean$\pm$std)~$\downarrow$}}\\ \cline{3-12}
            & &Liver &R. Kidney &L. Kidney &Spleen &Avg &Liver &R. Kidney &L. Kidney &Spleen &Avg. \\ 
            \hline
            \hline
            No Adaptation &- &49.2$\pm$9.6 &43.4$\pm$24.9 &65.4$\pm$18.1 &62.4$\pm$20.9 &55.1$\pm$14.3 &8.7$\pm$2.0 &10.8$\pm$4.6 &7.4$\pm$3.1 &4.5$\pm$2.9  &7.9$\pm$2.5 \\
            Supervised &- &93.5$\pm$1.2 &91.3$\pm$1.2 &92.1$\pm$2.6 &91.1$\pm$5.6 &92.0$\pm$2.7 &0.7$\pm$0.1 &0.6$\pm$0.1 &0.6$\pm$0.2 &0.5$\pm$0.2 &0.6$\pm$0.2 \\
            \hline
            SIFA~\cite{chen2020unsupervised} &\ding{56} &89.0$\pm$3.2 &83.8$\pm$4.0 &82.7$\pm$5.8 &84.6$\pm$8.0 &85.0$\pm$5.9 &1.2$\pm$0.5 &1.3$\pm$0.6 &1.5$\pm$0.7 &1.6$\pm$0.9 &1.4$\pm$0.8\\
            DAG-Net~\cite{xian2023unsupervised} &\ding{56} &84.8$\pm$4.6 &85.9$\pm$3.9 &86.7$\pm$3.6 &88.1$\pm$7.4 &86.4$\pm$4.9 &1.6$\pm$0.6 &1.1$\pm$0.5 &1.2$\pm$0.8 &0.9$\pm$0.7 &1.2$\pm$0.7 \\
            \hline
            DPL~\cite{chen2021source}&\ding{52}&70.1$\pm$6.9 &52.9$\pm$14.2 &65.7$\pm$12.5 &70.9$\pm$13.2 &64.9$\pm$8.8 &4.6$\pm$2.0 &7.9$\pm$3.1 &7.5$\pm$2.8 &3.2$\pm$2.6 &5.8$\pm$2.0 \\
            FSM~\cite{yang2022source}&\ding{52}&83.2$\pm$3.8 &74.5$\pm$4.5 &75.1$\pm$4.6 &76.2$\pm$9.8 &77.3$\pm$7.0 &2.8$\pm$0.8 &4.2$\pm$1.6 &5.0$\pm$1.9 &2.6$\pm$1.4 &3.7$\pm$1.3 \\
            AdaMI~\cite{bateson2022source}&\ding{52} &\textbf{90.2$\pm$1.0} &81.4$\pm$3.3 &82.6$\pm$4.7 &80.2$\pm$7.1 &83.6$\pm$6.6 &\textbf{1.0$\pm$0.5} &1.8$\pm$0.9 &1.6$\pm$1.0 &2.4$\pm$1.2 &1.7$\pm$0.9 \\
            Hong et al~\cite{hong2022source}&\ding{52}&88.1 &80.8 &\textbf{88.1} &79.2 &84.1 &- &- &- &- &-\\
            Ours &\ding{52} &89.9$\pm$2.7 &\textbf{84.5$\pm$6.8} &84.9$\pm$4.0 &\textbf{85.2$\pm$7.8} &\textbf{86.1$\pm$6.3} &1.3$\pm$0.5 &\textbf{1.4$\pm$0.4} &\textbf{1.3$\pm$0.6} &\textbf{1.2$\pm$0.8} &\textbf{1.4$\pm$0.7} \\
            
            \hline
            \hline

            \hline
            &\multicolumn{9}{c}{\textbf{Abdominal CT $\rightarrow$ Abdominal MRI}}\\
            \hline
            \multirow{2}{*}{Method} &\multirow{2}{*}{Source-free}  &\multicolumn{5}{c||}{\textbf{Dice (\%, mean$\pm$std)~$\uparrow$ }} &\multicolumn{5}{c}{\textbf{ASSD (voxel, mean$\pm$std)~$\downarrow$}}\\ \cline{3-12}
            & &Liver &R. Kidney &L. Kidney &Spleen &Avg &Liver &R. Kidney &L. Kidney &Spleen &Avg. \\ 
            \hline
            \hline
            No Adaptation &- &66.5$\pm$6.8 &81.6$\pm$14.6 &78.8$\pm$19.5 &70.1$\pm$11.6 &74.3$\pm$11.0 &3.5$\pm$1.6 &1.9$\pm$1.0 &3.0$\pm$1.6 &5.1$\pm$2.3  &3.4$\pm$2.1 \\
            Supervised &- &93.6$\pm$3.5 &94.2$\pm$1.8 &90.4$\pm$3.7 &92.2$\pm$2.6 &92.6$\pm$2.1 &0.5$\pm$0.2 &0.3$\pm$0.1 &0.4$\pm$0.1 &0.4$\pm$0.2 &0.4$\pm$0.2 \\
            \hline
            SIFA~\cite{chen2020unsupervised} &\ding{56} &87.1$\pm$4.6 &89.1$\pm$2.8 &84.2$\pm$3.9 &88.3$\pm$2.4 &87.2$\pm$3.8 &1.6$\pm$0.6 &0.8$\pm$0.2 &1.5$\pm$0.8 &1.7$\pm$0.9 &1.5$\pm$0.8\\
            DAG-Net~\cite{xian2023unsupervised} &\ding{56} &86.3$\pm$3.3 &89.0$\pm$2.0 &89.9$\pm$2.0 &90.6$\pm$2.9 &89.0$\pm$2.6 &1.8$\pm$0.6 &0.9$\pm$0.2 &0.9$\pm$0.4 &1.3$\pm$1.0 &1.2$\pm$0.6 \\
            \hline
            DPL~\cite{chen2021source}&\ding{52}&77.2$\pm$2.0 &83.5$\pm$12.5 &80.3$\pm$13.6 &83.3$\pm$7.2 &81.1$\pm$8.3 &3.0$\pm$0.6 &2.3$\pm$1.6 &2.5$\pm$1.4 &3.1$\pm$1.7 &2.7$\pm$1.4 \\
            FSM~\cite{yang2022source}&\ding{52}&84.3$\pm$3.1 &83.5$\pm$6.6 &82.1$\pm$9.0 &84.2$\pm$5.6 &83.5$\pm$4.3 &2.1$\pm$0.5 &1.5$\pm$0.6 &1.6$\pm$0.7 &2.2$\pm$0.6 &1.9$\pm$0.6 \\
            AdaMI~\cite{bateson2022source}&\ding{52} &85.5$\pm$2.4 &86.4$\pm$5.1 &82.1$\pm$7.6 &89.9$\pm$3.2 &86.0$\pm$3.1 &\textbf{1.9$\pm$0.4} &1.2$\pm$0.5 &2.2$\pm$0.6 &\textbf{1.3$\pm$0.7} &1.7$\pm$0.5 \\
            Hong et al~\cite{hong2022source}&\ding{52}&\textbf{88.4} &89.1 &86.4 &\textbf{91.1} &88.8 &- &- &- &- &-\\
            Ours &\ding{52} &86.1$\pm$0.5 &\textbf{91.7$\pm$5.1} &\textbf{88.6$\pm$8.0} &90.4$\pm$2.2 &\textbf{89.2$\pm$3.3} &2.0$\pm$0.3 &\textbf{0.7$\pm$0.2} &\textbf{1.0$\pm$0.4} &1.5$\pm$0.5 &\textbf{1.3$\pm$0.5} \\
            
            \hline
            \hline
        \end{tabular}

    }}
\label{tab1}
\end{table*} 

\begin{figure}[t]
\includegraphics[width=\textwidth]{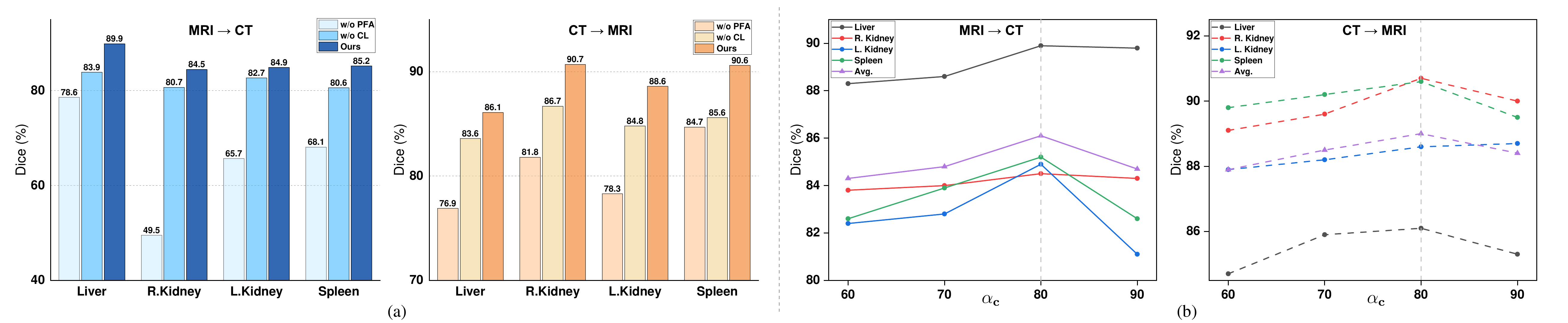}
\caption{(a) Ablation analysis of proposed two SFDA stages.  ``w/o CL'' denotes only the PFA is performed; ``w/o PFA'' denotes directly optimizing the contrastive loss according to the source model prediction. (b) Effect of different uncertainty percentile $\alpha_c$ on the adaptation performance.} \label{fig4}
\end{figure}

The quantitative evaluation results are presented in Table.\ref{tab1}. Compared to the upper and lower bounds in both directions, a huge performance gap can be observed due to the severe domain shifts between MRI and CT modalities. In ``MRI to CT'' direction, our method remarkably outperforms all other SFDA approaches on the right kidney and spleen, achieving the highest average Dice of 86.1\% and the lowest average ASSD of 1.4. Moreover, compared with recent UDA methods, our method obtains competitive results on average Dice and ASSD, which may be due to the use of unreliable predictions. As for ``CT to MRI'' direction, our method similarly shows great superiority on most organs as well, achieving the best performance in terms of both the average Dice (89.2\%) and ASSD (1.3) among all SFDA methods. Fig. \ref{fig3}(a) shows the segmentation results obtained by existing and our methods in both modalities. As observed, DPL is prone to amplify the initial noisy regions since it directly discards the unreliable pixels in self-training. For comparison, our method substantially rectificate the uncertain regions from the initial prediction, and details are shown in Fig. \ref{fig3}(b).  \\
\textit{\textbf{Ablation Study.}} In Fig. \ref{fig4}(a), we verify the effectiveness of the proposed two SFDA stages by removing each stage while keeping the other. The consecutive two stage adaptation leads to the best performance, while the drop in Dice is more significant if we remove the PFA stage. This result is not surprising because, without PFA, the source model prediction is too noisy to sample the qualified query and negative pixels for contrastative learning. We also study the impact of different uncertainty percentile $\alpha_c$ in Fig. \ref{fig4}(b). This parameter has a certain impact on performance, and we find $\alpha_c = 80\%$ achieves the best performance for most organs. Large $\alpha_c$ may introduce low-confidence query samples for supervision, and small $\alpha_c$ will drop some informative negative samples.

\section{Conclusion}
In this paper, we propose a novel two-stage framework to address the source-free domain adaptation problem in medical image segmentation. We first introduce a bi-directional transport cost to encourage the alignment between target features and source class prototypes in the prototype-anchored feature alignment stage. Also, a contrastive learning stage using unreliable predictions is further devised to learn a more compact target feature distribution. Sufficient experiments on the cross-modality abdominal multi-organ segmentation task validate the effectiveness and superiority of our method against other strong SFDA baselines, even some classical UDA approaches.
%
%
\bibliographystyle{splncs04}
\bibliography{refs}

\end{document}